\documentclass[12pt, twocolumn]{article}

\usepackage{amsmath}
\usepackage{graphicx}
\usepackage{url}
\usepackage{apacite}

\usepackage[a4paper, margin=1in]{geometry}

\title{The advantages of context-specific language models - the case of the Erasmian Language Model}
\author{João Gonçalves, Nick Jelicic, Michele Murgia, Evert Stamhuis\\
        \small Erasmus University Rotterdam\\
        \small \texttt{ferreiragoncalves@eshcc.eur.nl}}
\date{\today}

\begin{document}

\maketitle

\begin{abstract}
    The current trend to improve language model performance seems to be based on scaling up with the number of parameters (e.g. the state of the art GPT4 model has approximately 1.7 trillion parameters) or the amount of training data fed into the model. However, this comes at significant costs in terms of computational resources and energy costs that compromise the sustainability of AI solutions, as well as risks relating to privacy and misuse. In this paper we present the Erasmian Language Model (ELM) a small context-specific, 900 million parameter model, pre-trained and fine-tuned by and for Erasmus University Rotterdam. We show how the model performs adequately in a classroom context for essay writing, and how it achieves superior performance in subjects that are part of its context. This has implications for a wide range of institutions and organizations, showing that context-specific language models may be a viable alternative for resource constrained, privacy sensitive use cases.
\end{abstract}

\section{Introduction}
Large Language Models (LLMs) have been one of the most visible machine learning applications of the past decade, attracting a large amount of academic research and industry investment in search of higher performance and productivity gains. The dominant trend in improving language models seems to be based on scaling up the number of parameters (e.g. GPT4 has approximately 1.7 trillion parameters) or the amount of training data fed into the model (e.g. the Common Crawl dataset currently has more than 75 TB of textual data). However, this scaling usually comes at significant costs in terms of computational resources, which translate to energy \cite{strubell_energy}, financial, and environmental costs, and data collection, which in turn raise concerns regarding legal, privacy, quality, and responsibility \cite{palacios2023intersection} trade-offs.

While there has been progress in increasing the efficiency of training and inference of language models through model architecture improvements \cite{compute_optimal}, and processes such as quantization \cite{quantization} and distillation \cite{sanh2019distilbert}, these do not address the overall trend of an exponential growth in terms of use of resources. Furthermore, normalizing a high volume of computational resources for model training also raises issues in terms of AI governance, where the advanced capabilities of LLMs will be concentrated in the few select organizations that can acquire a large number of GPUs required for model training and deployment.

In this paper, we explore an alternative path to LLM adoption, harnessing a small, tailored and community driven language model that only excels at the tasks it was designed to do. We call this a context-specific model, where training data, model size, fine-tuning data, testing and deployment are guided by the context that the model is created for. With this context-specific model, we aim to address limitations of language models related to efficiency, use of resources, privacy and ethics.

Considering recent findings on how data quality is a key component of LLM performance \cite{gunasekar2023textbooks}, we pre-train the Erasmian Language Model (ELM) a 900M parameter Llama-2 based language model exclusively on the data of Erasmus University Rotterdam, a large Dutch higher education institution. We then fine-tune an instruct model with the Alpaca dataset, in English and in Dutch. We show that, despite having a much more limited number of parameters and training data, this model specializes in the subjects taught and researched at Erasmus University, and is suitable for classroom activities. The model produces credible text in the domains of Social Sciences, Humanities, and Medicine, but fails at any other topic, including simple reasoning tasks. As a byproduct of data and usage control, we also show that the model does not require the standard guardrails using reinforcement learning with human-feedback.

\section{Related Work}

\subsection{Improving language model efficiency}
The performance of LLMs is usually accessed by means of standardized general benchmarks such as Massive Multitask Language Understanding (MMLU) \cite{hendrycks2021measuringmassivemultitasklanguage}, where large players in the field compete to top others by expanding the number of parameters \cite{achiam2023gpt}, number of tokens in the training data \cite{zhang2024tinyllamaopensourcesmalllanguage}, improving model architecture \cite{de2024griffinmixinggatedlinear}, among other strategies to increase performance. However, the environmental, resource, and privacy implications of a pursuit of scale in LLMs have been increasingly highlighted as relevant values in addition to performance \cite{tokayev2023ethical}, shifting the focus to model efficiency and trustworthiness.

An important part of the search for efficiency gains has been to find compute optimal calculations for model size and number of tokens, that aim to ensure LLM are not overparameterized \cite{compute_optimal}. While this does not necessarily address the pursuit of scale on its own, it does make it so that model training is designed in a more balanced and efficient way. From the inference side, techniques like quantization \cite{quantization}, representing model weights at a lower precision, or knowledge distillation \cite{buciluǎ2006model}, training smaller models based on larger models, have shown promise in decreasing the computational and energy costs of model deployment. These efforts aim to improve the efficiency of model training or inference, but do not directly address the need for larger models.

A second strand of LLM research has focused on training smaller models that match or outperform larger language models. One high profile effort along this line has been Microsoft's training of its Phi family of models \cite{gunasekar2023textbooks}. The assumption for these models is that data quality is more important than data quantity or model size to determine the quality of outputs. However, when we examine Phi-3 \cite{abdin2024phi3technicalreporthighly}, the latest iteration of the Phi models, we see that while a small model is technically trained, the actual size of the training data surpasses 3 trillion tokens, which to some extent defeats the efficiency gains of a smaller model by still imposing high training compute requirements on the system.

\subsection{context-specific language models}
An alternative in pursuing efficiency gains with large language models is to abandon their general purpose nature in order to train context-specific models, that still excel at a wide range of tasks within that context. The underlying assumption behind the performance of large language models is that, through transfer learning \cite{pan2009survey} and fine-tuning, large language models and their fine tuned versions will always outperform specifically trained models \cite{howard2018universallanguagemodelfinetuning}. However, when we consider performance also accounting for environmental, privacy and other costs, this general purpose pre-training and domain specific fine-tuning becomes a less attractive option.

One of the most popular domains where context-specific models emerge is in programming code. Some of these models \cite{li2023starcodersourceyou} are pre-trained and fine-tuned only on code and, while they still possess  a large number of parameters and training data, they mitigate some of the computational costs that general purpose models have by restricting the training dataset. These code models, however, still have copyright and privacy implications that stem from leveraging the work of GitHub contributors, raising ethical concerns about using user-contributed data.

A second strand of context-specific models leverage organization specific data to pretrain custom language models. A notable example is a language model trained by financial information provider Bloomberg on a mixture of their own data and general data \cite{wu2023bloomberggptlargelanguagemodel}. For this language model Bloomberg validated the model not only in general purpose benchmarks such as the ones mentioned above, but also in terms of performance for internal tasks. This model, however, still used 50\% of data that was external to the company.

\subsection{Epistemic-functional approach}
The decision to pursue context-specific language models is not a purely technical one, it also represents a shift in the frameworks and values under which LLMs are developed and used. Our approach to context-specific models focuses on their epistemic functionality. It looks at how the use of LLMs within knowledge processes of a given context work and make sense to those involved. They thus have to fit the epistemic needs, wants and capacities of a community. Yet this prerequisite is never isolated from questions of power and so one has to take the values and norms of the community into account \cite{russo2024connecting}. 'The public' as community is often related to values such as care for the environment, privacy and human dignity. Whereas a corporation, notwithstanding overlap with other communal values, will have distinctly other values, such as profit.

What counts as functional is in sum strongly intertwined with the sources of normativity that underpin (epistemic) communities. So if an LLM is trained for an academic community, as is the case here, its involvement is a necessity. Especially because norms are objects of contestation - one cannot treat a community as an epistemic entity, nor purpose-specific models as functional, when norms are taken to be static. An epistemic-functional approach is thus as much concerned with LLMs as objects as it is as processes. In this regard we see our approach as an application of AI realism, which focuses on non-moral (epistemic) normativity to design, develop and evaluate AI technologies \cite{murgia2024overcoming}.

Our approach also bears similarities with the more familiar value sensitive design (VSD). VSD aims to explicitly integrate values into the design of technologies, lest it runs the risk of becoming undesirable \cite{friedman2007human}. Like our approach, VSD emphasises the context of technology and the salience of respective values in that context. The upshot is indeed relevance within contextual bounds; the cost is reducing normativity to stakeholder preferences \cite{jacobs2021value}. Put differently, those values and norms risk becoming entirely and solely relative to the context in which they are employed. The consequences are: (i) the lack of an independent justificatory standard by which the contestation of values and norms can be guided, and (ii) the increased narrowness of values as contexts within contexts unfold, ironically rendering them context insensitive, powerless and difficult to operationalise.

The epistemic-functional approach differs from VSD insofar as it links functionality to knowledge relations. While those relations are inscribed into communities and thus their values, they are not reduced to them. The epistemic-functional approach looks, again, at the epistemic needs, wants and capacities of those communities. This vantage point invites a standard by which communal values can be evaluated and judged, even though the two are intertwined. Key here is involving the epistemic community so that the standard remains dynamic and avoids the mire of context exhaustion.

Finally, our approach does call for decisionism. One has to decide, after all, which public values correspond with which community. If the epistemic community must be involved in the development of an LLM, it cannot be the case that we as researchers observe this involvement as isolated observers. In fact, such a development does not begin until a party, implicitly or explicitly, decides on the values underpinning it. With this in mind we propose that the values of an academic community largely overlap with what is generally known as public values. This stands in contrast to the LLMs that academic communities largely use, namely those developed and provided, that is: sold, by (big-)tech companies. Instead of subservience to these corporate-driven technologies, we propose to bring AI into the fold of academic politics.

\section{Methods}
To address the limitations of related work, we propose to train a language model, the Erasmian Language Model (ELM), on data that originated fully from a middle sized organization such as Erasmus University Rotterdam, validating it for the specific purposes of this institution, namely teaching and research. This offers a methodological template for organizations that face resource or privacy constraints in implementing language models.

\subsection{Model architecture}
ELM was based on the state-of-the-art architecture at the time, LLaMA 2  \cite{touvron2023llama}. The number of hidden layers and attention heads was scaled down linearly to train two versions of the model, a small model with 160 million parameters and a larger one with 900 million parameters, both in single precision 32-bit floating point. The small model was intended as a teaching tool, allowing students to fine-tune or train some custom versions of the model in their own machines. The language model was trained to serve as a broader purpose model to support research and education in tasks such as idea generation, essay writing, or assignment feedback.

We also trained our own version of the LLaMA 2 tokenizer. We follow LLaMa 2 and use SentencePiece to map tokens to a vocabulary size of 32K. This was aimed to keep the vocabulary size to the strictly necessary so that tokens from languages irrelevant to our data were not considered, but also to ensure that the domain specific performance of the model was not compromised by an inadequate tokenizer. 

\subsection{Datasets}
For pretraining, we compile a corpus of documents strictly produced by Erasmus University Rotterdam, to ensure that our model reflects the institution and its contexts. We collect data from two sources: the EUR research output, and the Erasmus University Thesis repository. The total EUR corpus consists of roughly 2.7B tokens and spans from 1969 to 2023. This dataset is particularly useful for Erasmus University because it represents the knowledge of its academics through the Research Output corpus, and the knowledge of its students through the Master Theses corpus. This corpus cannot be made publicly available because of copyright restrictions, the custom made instruction and reinforcement learning datasets are available on GitHub \url{https://github.com/Joaoffg/ELM}.  

\begin{figure}
    \centering
    \includegraphics[width=1\linewidth]{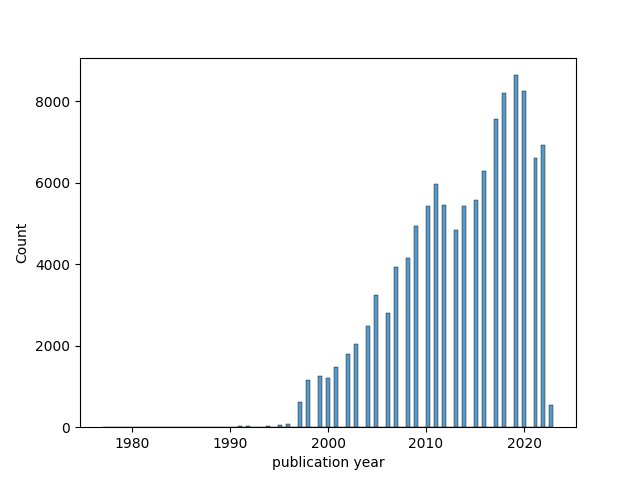}
    \caption{Research papers per year (research output)}
    \label{fig:enter-label}
\end{figure}

\subsubsection{EUR Research output}
We extracted all DOIs from the EUR CRIS system (Pure) and collected the PDFs. This resulted in 75.355 documents from 1977 to 2023. Figure 1 shows the number of documents per year. The research output consists of 1.917.100.206 tokens. Documents are written mostly in English and in Dutch.

\subsubsection{EUR Theses}
The Erasmus University has its own thesis repository. We collected 16.983 theses from the repository that do not have any restrictions to usage from 1969 to 2022. Figure 2 shows the number of theses per year. The EUR Thesis dataset consists of 864.152.740 tokens. Documents are written mostly in English and in Dutch. Only public theses were used. Theses that could contain sensitive information, such as confidential business data, were removed from the dataset to preserve privacy.

\begin{figure}
    \centering
    \includegraphics[width=1\linewidth]{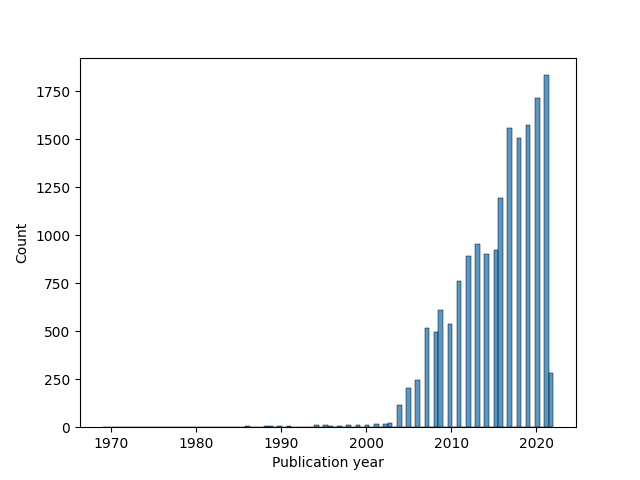}
    \caption{Number of theses per year}
    \label{fig:enter-label}
\end{figure}

\subsubsection{Instruction dataset}
For further instruction tuning of the foundation model we constructed a dataset consisting of 485 input-output pairs. These input pairs were constructed by student assistants, who were asked to write examples representing how they would like the model to be used, they were paid the standard rates in accordance with the Dutch Collective Labor Agreement for Universities. We convert the instruction pairs to match the alpaca instruction template. 

While the generated pairs were high quality, they proved insufficient for the model to properly learn how to follow instructions. Therefore, they were supplemented with the cleaned version of the Alpaca dataset \cite{alpaca} and a version of it translated into Dutch \cite{vanroy2023language}.

\subsubsection{Reinforcement learning}
We constructed a dataset for direct preference optimization \cite{rafailov2023directpreferenceoptimizationlanguage} with the same student assistants who took part in the instruction dataset generation. They were simply instructed to choose the best output out of two generated by the model according to their judgement. Unlike other models, we chose not to impose guardrails for the type of language being generated because: 1) Our pretraining is limited to the type of language and documents we wish to produce, thus eliminating the need for post-training filtering; 2) In certain academic domains, such as history, students and scholars have to engage with problematic and outdated language, therefore adjusting the model to avoid usage of this language would not be desirable.

\subsection{Training}
We first experimented with fine-tuning Meta's Llama 2 model based on Erasmus University data to avoid the computational costs of the pretraining stage. However, the model performed inadequately for academic purposes, resorting to general information more frequently than domain specific information required for Erasmus University.

Pretraining of ELM-small was conducted on a single Nvidia A10 GPU for 3 epochs, taking a total of 48 hours, using the datasets and transformers libraries from HuggingFace. Pretraining of ELM-medium was also conducted on a single Nvidia A10 GPU for 3 epochs, taking a total of 720 hours (30 days) of training. Training was conducted on the secure SURF research cloud provided to Dutch universities, showing that training of context-specific models is feasible with limited access to resources.

Fine-tuning of ELM-medium happened in the same A10 GPU, using low rank adaptation (LoRA) to fine tune the model as an instruct version of ELM. This version was then further trained with reinforcement learning with human feedback (RLHF) using direct preference optimization with the Huggingface's TRL library.

To verify that the model aligned with the goals and values of the community it was intended to serve, Erasmus University Rotterdam, ELM-small was presented in a public event where participants could interact with the model. This ensured that any interested parties had an opportunity to voice concerns and expectations regarding the model before the larger model was trained.

\section{Results}
ELM was trained exclusively for the purposes of Erasmus University Rotterdam, which makes assessing its performance through traditional benchmarks such as MMLU less relevant for the task at hand. Therefore, in addition to an MMLU assessment, we have also conducted a qualitative assessment of the model with its student end users.

\subsection{Qualitative assessment}
Qualitative assessment of ELM took place in the scope of a bachelor minor on the topic of artificial intelligence where students were tasked to write an essay with the support of the language model. They were then asked to reflect on the model, focusing on its advantages and disadvantages. The version under assessment was ELM-small, to facilitate scaling up local deployment of the model (100 students enrolled in the minor). To prevent any social desirability bias in the evaluation caused by the fact that the main developer of the model was also the teacher assessing student submissions, the assignment was embedded in the course but was not graded. After submitting the assignment, students were asked if they would like their assessment to be part of the research paper. 23 students consented to this.

Students considered that the outputs of the model were coherent and were very aligned with academic language. However, they emphasised that the model had a tendency to go off topic at times, potentially caused by the boundaries of the training data coverage. \textit{"Although the output that was generated did not answer the question per se, the output is coherent and easy to read."}

Students also identified a limitation of the model in terms of its ability to generate coherent long texts. This shows that our decision to truncate text chunks to 256 tokens limits to some extent the model's applicability to academic purposes, where longer texts are typically the norm. When confronted with more abstracts questions, students also noted a tendency for it to make up words or references, losing coherence and sentence structure. The shortcomings and advantages of ELM-small are well summarized by a student, who mentioned: \textit{"the text ELM produced has little to do with a well written, cohesive essay. [...] Yet, when compared to other LLMs if felt as if ELM emulated my writing style much more accurately, possibly due to our shared source of education".}

In summary, students had different experiences with ELM in terms of coherence and language quality, depending on how closely the used prompts related to Erasmus University topics, and the generation parameters they specified, particularly \textit{max length}, \textit{repetition penalty}, and \textit{temperature}. However, there was consensus in recognizing the language as distinctly academic, and also closer to the student's own writing.

\subsection{Quantitative assessment}

As mentioned above, general purpose benchmarks are less suitable to assess the performance of ELM due to its domain and context-specific nature. However, we can resort to multidisciplinary benchmarks such Massive Multitask Language Understanding (MMLU) \cite{hendrycks2021measuringmassivemultitasklanguage} to verify if the model displays the expected behavior, i.e. specializing in the disciplines covered by Erasmus University: social sciences, humanities, and medicine. 

To run the MMLU benchmark on ELM, we fine-tuned the ELM-medium model on the training set of the MMLU benchmark for 1 epoch using LoRA. The benchmark has a baseline performance of 0.25, given that each multiple choice question has 4 answer options. However, even after the fine-tuning, the model did not always output answers within the specified options of 0-3, causing values to sometimes be below 0.25. However, as mentioned previously, our goal with using the MMLU benchmark is not to use it as a comparison term to other models, but to verify that it is specializing in the domains that are covered by Erasmus University Rotterdam. 

The MMLU benchmark is comprised of 57 different subjects, ranging from abstract algebra to world religions. The benchmark creators themselves assigned broader container fields to these subjects, namely STEM, Social Sciences, Humanities and Other. Given that the subject coverage at Erasmus University does not linearly overlap with these categories, we also conducted our own subject categorization, labelling subjects as EUR or non-EUR. A detailed table with the performence per subject, and the corresponding categorization, can be found at the end of this document in Table 1.

\begin{figure*}
    \centering
    \includegraphics[width=\linewidth]{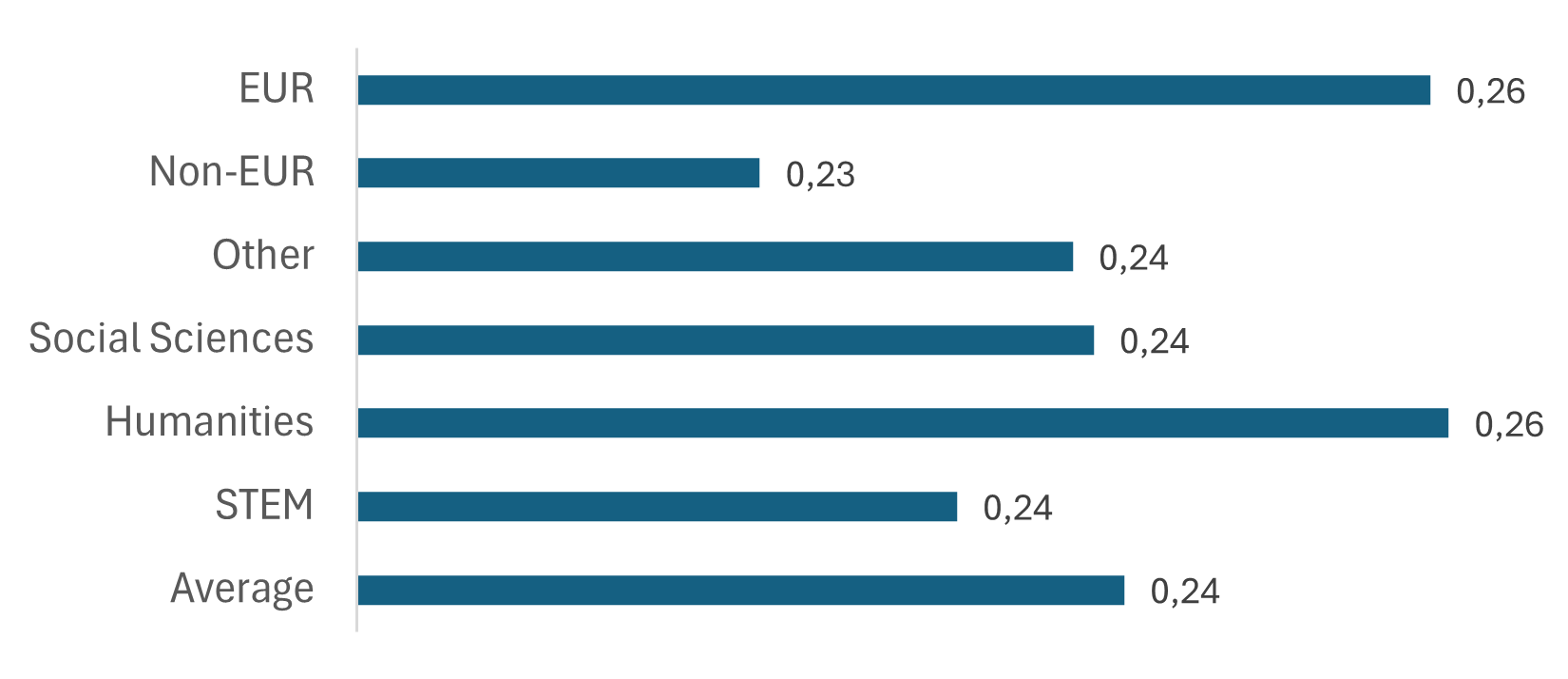}
    \caption{Performance of ELM per subject according to MMLU benchmark}
    \label{fig:ELM_MMLU}
\end{figure*}

Fig. \ref{fig:ELM_MMLU} shows that the performance of ELM in the overall benchmark is well below state of the art models, but this value is less relevant for our assessment given that the benchmark itself minimally overlaps with EUR research. The relative performance between subjects, however, shows that the model specializes in topics related to the social sciences and especially the humanities, in line with the profile of Erasmus university. When we compare the subjects that are covered by EUR with non-EUR subjects, the difference is even clearer, with EUR subjects scoring .26 accuracy and non EUR subjects having the lowest score at .23. Despite not being large, this difference showcases how the model is specializing in subjects covered by EUR, achieving its intended behavior by adjusting to the context it was trained for.

\subsection{COMPASS assessment}
\begin{figure}
    \centering
    \includegraphics[width=1\linewidth]{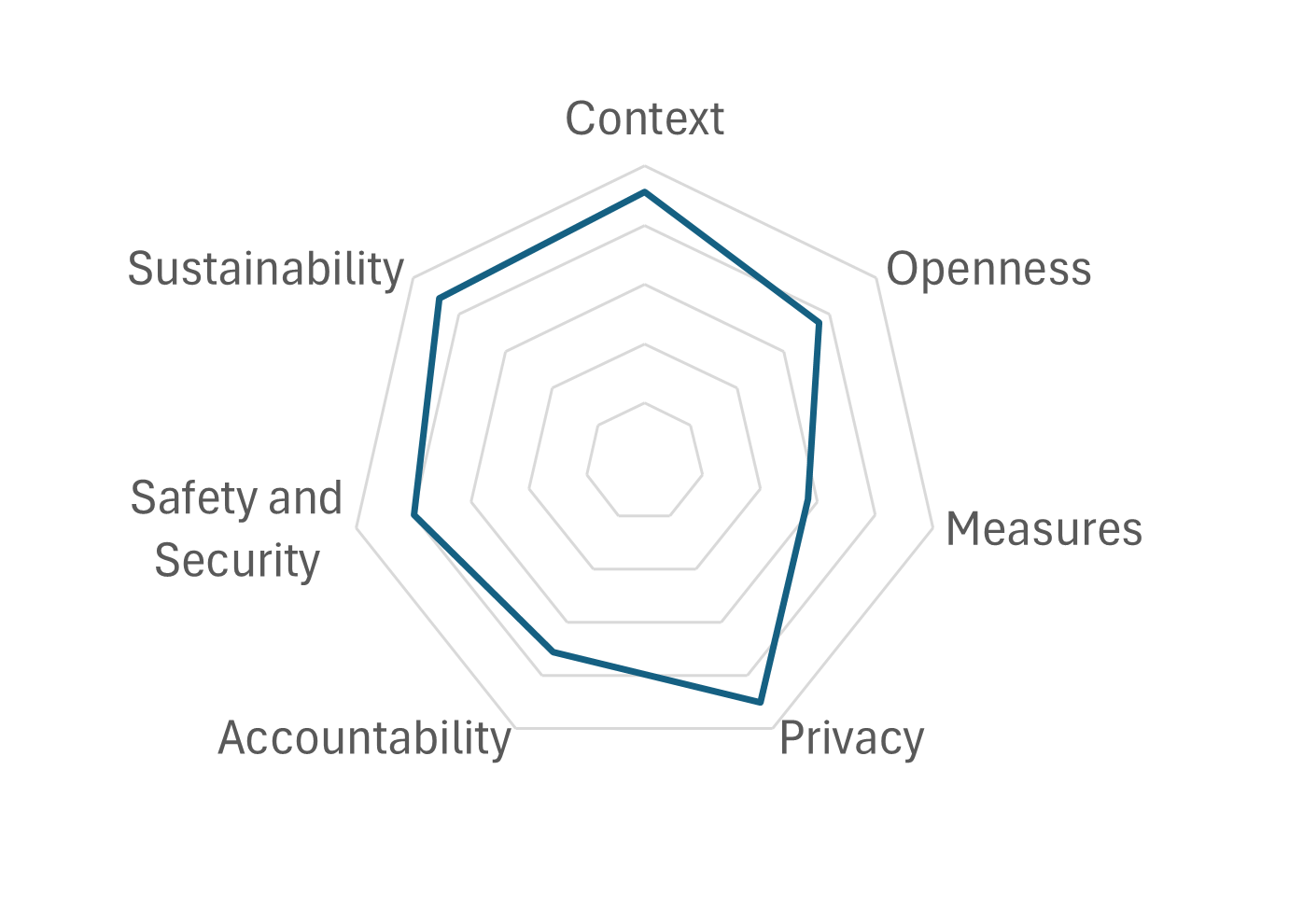}
    \caption{COMPASS Assessment}
    \label{fig:compass-asssessment}
\end{figure}

To assess to what extent the ELM represents an improvement over other language models in terms of trustworthiness we used the COMPASS framework, developed under the H2020 EU SPATIAL project \url{https://spatial-h2020.eu/compass-questionnaire/}. COMPASS is a self-reflective AI evaluation tool that allows developers to situate their systems in terms of trustworthiness at several points of the development process under 7 criteria. The COMPASS assessment as of July 2024 is presented in Fig \ref{fig:compass-asssessment} and available at\url{https://github.com/Joaoffg/ELM/blob/main/The-Compass-Questionnaire_ELM_2024_07_03.pdf}.

As expected and intended, ELM scores highly on most of the criteria, especially in terms of context definition, privacy and sustainability. However, the assessment tool also allows us to identify key areas of improvement for the model, namely in terms of more clearly defining measures and metrics for success and failure, and also embedding explainability methods in the deployment of the model.

\section{Conclusion}
The ELM experiment aimed to show that, by focusing on domain and context-specific data, acceptable performance could be achieved with language models even in the presence of significant resource constraints. The model was trained and deployed on a single A10 GPU, and all project costs related to the version presented in this paper amount to less than 4000 euros. When compared to large commercial models of companies that have a magnitude of resources that is much larger, ELM is inferior in many aspects. However, when looking at the qualitative and quantitative performance of ELM in relation to its context of application, our findings show that even with modest resources the model already delivers useful performance in essay writing, and in acquiring some level of specialized knowledge.

The ELM prototype also showed the benefits of an iterative development process, where usage and feedback from students in a classroom context showed, for instance, that increasing the length of the text chunks fed to the model was a crucial requirement for academic applications. The flexible nature of small, context-bound, language models means that it is easier to iterate through different versions of the model based on feedback, and that major issues can be addressed at a model training stage. This mitigates the need for corrective measures after deployment, such as the editing of prompts to ensure greater diversity of outputs \cite{kirk2023understanding}. A well defined context of use facilitates the alignment of LLM development with end-user expectations.

ELM is a proof of concept that shows that organizations with limited resources, but high requirements in terms of efficiency, privacy, and trustworthiness, should look into context-specific language models as a viable alternative to large commercially driven language models. It is important to note that ELM itself was not intended to be more than this proof of concept, with a deployment version of the model requiring further custom development by, for instance, relying less on the alpaca dataset for fine-tuning. However, even at its proof of concept stage, ELM has already proven its usefulness in classroom and research environments within Erasmus University Rotterdam, fulfilling its purpose as a context-specific model. Its smaller size and the traceability of its training data also mean that many of the privacy and impact concerns that affect other language models do not pose themselves in the context of ELM, showcasing the some of the advantages of its context-specific approach in addition to performance.

\section{Acknowledgments}
The authors would like to acknowledge the contributions of Ilias McAuliffe and Marios Papamanolis in producing the instruct and reinforcement learning datasets, as well as providing insights to model development. We also acknowledge the funding of ELM by Convergence AI, Data \& Digitalisation and Erasmus Trustfonds, and the logistical support of the Erasmus Center for Data Analytics, SURF, and the Erasmus Future Library Lab. We also thank the students of the AI and Societal Impact minor for their role in testing and providing critical feedback to the model.

\section{Author contributions}
João Gonçalves coordinated the technical development of ELM and wrote the majority of the paper. Nick Jelicic developed a substantial part of the code and acquired the data. Michele Murgia was the overall project lead, wrote the section on the epistemic-functional approach and managed student assistants. Evert Stamhuis provided insights to the project and facilitated collaboration and coordination.

\bibliographystyle{apacite}{}
\bibliography{references.bib}

\clearpage

\begin{table}[]
\label{tab:ELM_MMLU_performance}
\caption{Performance of ELM per subject}
\footnotesize
\begin{tabular}{llll}
                                        & Accuracy & Subject         & EUR     \\
abstract\_algebra                       & 0,25     & STEM            & Non-EUR \\
anatomy                                 & 0,16     & STEM            & Non-EUR \\
astronomy                               & 0,17     & STEM            & Non-EUR \\
business\_ethics                        & 0,33     & Other           & EUR     \\
clinical\_knowledge                     & 0,25     & Other           & EUR     \\
college\_biology                        & 0,3      & STEM            & Non-EUR \\
college\_chemistry                      & 0,26     & STEM            & Non-EUR \\
college\_computer\_science              & 0,24     & STEM            & Non-EUR \\
college\_mathematics                    & 0,14     & STEM            & Non-EUR \\
college\_medicine                       & 0,24     & Other           & EUR     \\
college\_physics                        & 0,25     & STEM            & Non-EUR \\
computer\_security                      & 0,25     & STEM            & Non-EUR \\
conceptual\_physics                     & 0,29     & STEM            & Non-EUR \\
econometrics                            & 0,24     & Social Sciences & EUR     \\
electrical\_engineering                 & 0,26     & STEM            & Non-EUR \\
elementary\_mathematics                 & 0,25     & STEM            & Non-EUR \\
formal\_logic                           & 0,29     & Humanities     & EUR     \\
global\_facts                           & 0,19     & Other           & Non-EUR \\
high\_school\_biology                   & 0,19     & STEM            & Non-EUR \\
high\_school\_chemistry                 & 0,24     & STEM            & Non-EUR \\
high\_school\_computer\_science         & 0,27     & STEM            & Non-EUR \\
high\_school\_european\_history         & 0,23     & Humanities     & EUR     \\
high\_school\_geography                 & 0,18     & Social Sciences & Non-EUR \\
high\_school\_government\_and\_politics & 0,28     & Social Sciences & EUR     \\
high\_school\_macroeconomics            & 0,27     & Social Sciences & EUR     \\
high\_school\_mathematics               & 0,23     & STEM            & Non-EUR \\
high\_school\_microeconomics            & 0,31     & Social Sciences & EUR     \\
high\_school\_physics                   & 0,23     & STEM            & Non-EUR \\
high\_school\_psychology                & 0,23     & Social Sciences & EUR     \\
high\_school\_statistics                & 0,24     & STEM            & Non-EUR \\
high\_school\_us\_history               & 0,28     & Humanities     & EUR     \\
high\_school\_world\_history            & 0,29     & Humanities     & EUR     \\
human\_aging                            & 0,3      & Other           & EUR     \\
human\_sexuality                        & 0,28     & Social Sciences & EUR     \\
international\_law                      & 0,19     & Humanities     & EUR     \\
jurisprudence                           & 0,22     & Humanities     & EUR     \\
logical\_fallacies                      & 0,29     & Humanities     & EUR     \\
machine\_learning                       & 0,26     & STEM            & Non-EUR \\
management                              & 0,17     & Other           & EUR     \\
marketing                               & 0,22     & Other           & EUR     \\
medical\_genetics                       & 0,23     & Other           & EUR     \\
miscellaneous                           & 0,26     & Other           & Non-EUR \\
moral\_disputes                         & 0,26     & Humanities     & EUR     \\
moral\_scenarios                        & 0,24     & Humanities     & EUR     \\
nutrition                               & 0,17     & Other           & Non-EUR \\
philosophy                              & 0,22     & Humanities     & EUR     \\
prehistory                              & 0,25     & Humanities     & EUR     \\
professional\_accounting                & 0,28     & Other           & EUR     \\
professional\_law                       & 0,26     & Humanities     & EUR     \\
professional\_medicine                  & 0,25     & Other           & EUR     \\
professional\_psychology                & 0,29     & Social Sciences & EUR     \\
public\_relations                       & 0,32     & Social Sciences & EUR     \\
security\_studies                       & 0,19     & Social Sciences & Non-EUR \\
sociology                               & 0,19     & Social Sciences & EUR     \\
us\_foreign\_policy                     & 0,12     & Social Sciences & Non-EUR \\
virology                                & 0,24     & Other           & EUR     \\
world\_religions                        & 0,32     & Humanities     & Non-EUR
\end{tabular}
\end{table}

\end{document}